\documentclass[letterpaper, 10 pt, conference]{format/ieeeconf}  

\IEEEoverridecommandlockouts                              

\usepackage{amsmath} 
\usepackage{amssymb}  
\usepackage{xcolor}
\usepackage[ruled,vlined,linesnumbered]{algorithm2e}
\usepackage{graphicx}
\usepackage{lipsum}
\usepackage[noadjust]{cite}
\usepackage{subcaption}
\usepackage{float}
\usepackage{wrapfig}
\usepackage{pifont}

\usepackage{hyperref}       
\usepackage{xurl}            
\usepackage{afterpage}

\title{\bf Predicting the Efficiency of $\text{CO}_{\text{2}}$ Sequestering by Metal Organic Frameworks Through Machine Learning Analysis of Structural and Electronic Properties}

\author{Mahati Manda \\
\texttt{mahathi.manda@gmail.com} \\
\texttt{Basis Independent Silicon Valley}}

\begin{document}

\maketitle

\thispagestyle{empty}
\pagestyle{empty}

\begin{abstract}
Due the alarming rate of climate change, the implementation of efficient $\text{CO}_{\text{2}}$ capture has become crucial. This project aims to create an algorithm that predicts the uptake of $\text{CO}_{\text{2}}$ adsorbing Metal-Organic Frameworks (MOFs) by using Machine Learning. These values will in turn gauge the efficiency of these MOFs and provide scientists who are looking to maximize the uptake a way to know whether or not the MOF is worth synthesizing. This algorithm will save resources such as time and equipment as scientists will be able to disregard hypothetical MOFs with low efficiencies. In addition, this paper will also highlight the most important features within the data set. This research will contribute to enable the rapid synthesis of $\text{CO}_{\text{2}}$ adsorbing MOFs.
\end{abstract}

\section{Introduction}
\label{sec:introduction}

Due to industrialization and over consumption, climate change has become an increasingly important issue to solve as the environment continues to deteriorate. Negative effects such as extinction, desertification, and the increase of natural disasters have made it impossible for the scientific community to ignore this phenomenon. The buildup of harmful and toxic substances, as well as greenhouse gasses in the air, has caused the destruction of the ozone layer and contributed to warming of the earth. In fact, a whopping 13\% of global childhood asthma cases have been attributed to excessive air pollution. 
Carbon dioxide is a notorious pollutant and irritant. It is emitted by cars, factories, and many more facilities incorporated into everyday modern lives. In order to decrease these carbon dioxide emissions, factories have implemented more efficient practices, solar power alternatives have risen in opposition to coal power, and electric cars have been designed. The creation of Carbon-Dioxide Adsorbing Metal-Organic Frameworks has proven to be a major contributor to the reduction methods applied to carbon dioxide and carbon dioxide-capturing technology. \cite{deng_yang_li_liang_shi_qiao_2020}  

\section{Metal-Organic Frameworks}
\label{subsec:MOF}
Metal-Organic Frameworks (MOFs) are crystalline porous materials made of linkers and metal ions. MOFs are akin to sponges in that they are structured to be porous. Due to these pores, MOFs can have high porosity and large surface areas. Furthermore, the topological diversity of MOFs can allow researchers to control many different features during the synthesizing process. For example, varying functional groups can cause MOFs to have different properties and results for the functions that they are designed to perform. MOFs are used for a plethora of applications such as gas storage, catalysis, and drug generation. Within the realm of gas storage, MOFs have been utilized to store and adsorb different types of greenhouse gasses such as carbon dioxide, methane, and nitrogen. In this paper, we will specifically consider MOFs that adsorb carbon dioxide. \cite{zhou_long_yaghi_2012}

\section{$\text{CO}_{\text{2}}$ Adsorbing MOFs}
\label{subsec:CO2AdsorbingMOFs}
The first carbon-capturing MOF was reported in 1998 and was dubbed MOF-2. Since then, more MOFs have been synthesized. There are 3 main types of $\text{CO}_{\text{2}}$ capture: pre-combustion capture, post-combustion capture, and oxy-fuel combustion capture. \cite{siegelman_mcdonald_gonzalez_martell_milner_mason_berger_bhown_long_2017} For this paper, we will exclusively discuss post-combustion $\text{CO}_{\text{2}}$ capture utilizing MOFs as the dataset includes only post-combustion MOF uptakes. Post-combustion flue gas is gas from power plants which went through $\text{CO}_{\text{2}}$ combustion and contains 25\% $\text{CO}_{\text{2}}$ and 75\% $\text{N}_{\text{2}}$. Since the $\text{CO}_{\text{2}}$ goes through combustion, the temperature of the flue gas starts out being very high. In order to continue with the capture, this flue gas is first cooled. Next, the flue gas goes through pre-treatment, in which impurities are removed. Finally, the $\text{CO}_{\text{2}}$ binds to the adsorption sites (adsorbophores) on the MOF. As more $\text{CO}_{\text{2}}$ starts binding, some $\text{CO}_{\text{2}}$ gets pushed out of the MOF, making it more unstable and inefficient at holding $\text{CO}_{\text{2}}$. At a certain point, the MOF reaches full capacity for storage. In order to determine whether a MOF is efficient and stable, $\text{CO}_{\text{2}}$ uptake is important to observe. \cite{sumida_rogow_mason_mcdonald_bloch_herm_bae_long_2011}
\section{Machine Learning}
\label{subsec:ML}
MOF generation can be expensive and time-inefficient, especially when the synthesized MOFs result in poor CO2 adsorption. Machine Learning can be used to display the most efficient MOFs and their properties, so that scientists may save resources. A subset of Artificial Intelligence, Machine Learning is used to understand and create models to fit different data structures. One sub-category of Machine Learning is regression, which is used to analyze the relationship between the independent and dependent features. ~\cite{tagliaferri_2017} When creating a regression line, considering 2 or 3 features may be possible, yet taking into account multiple features can increase the likelihood of errors. This task becomes more difficult depending on the size of the dataset. Thus, Machine Learning models can be used to standardize, apply, and compare different models in order to create a regression line in consideration of multiple features with the least mean standard error. In Machine Learning, the computer is training and testing the data to find relationships and trends. Within Machine Learning, it is important to identify whether to use supervised or unsupervised learning to determine the model. The difference between the two is that supervised learning involves the dataset to provide values for the dependent and independent variable whereas the unsupervised model expects solely values for the dependent features. In this paper, supervised learning is utilized. \cite{rahimi_moosavi_smit_hatton_2021}
\section{DATASET}
\label{subsec:Dataset}
For this research, a hypothetical database containing 324,426 $\text{CO}_{\text{2}}$-adsorbing MOFs has been used. For more details about this database, please visit: \url{https://archive.materialscloud.org/record/2018.0016/v3} A hypothetical dataset contains hypothetical MOFs or virtual MOFs created and tested by other Artificial Intelligence programs, which are yet to be synthesized. The dataset structure includes 324,426 rows and 42 columns. This corresponds to the 324,426 MOFs and 42 features present before cleaning the data. This dataset was used rather than Cambridge's CORE MOF Structural Dataset due to time limitations that would arise from analyzing CIF files through Machine Learning analysis. 
\section{DATA ANALYSIS AND PRE-PROCESSING}
\label{sec:DataAnalysisandProcessing}
As a part of data preparation, it is important to analyze prominent trends. A plethora of the graphs provided by the data analysis were weak and scattered, although some clear directions are present. The Carbon Dioxide Uptake is influenced significantly by factors such as the surface area, weight, void fraction, and void volume. (Figure 1).\cite{jablonka_ongari_moosavi_smit_2020}


\vspace{-0.05in}
\begin{figure}[ht!]
    \centering
    \includegraphics[width=\linewidth]{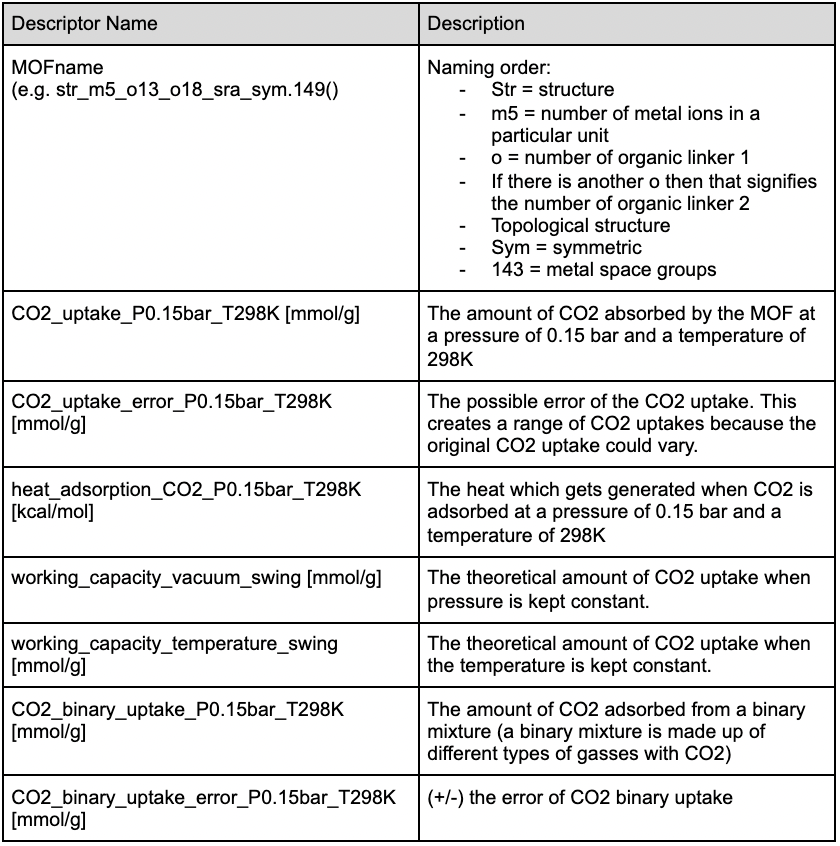}
    \includegraphics[width=\linewidth]{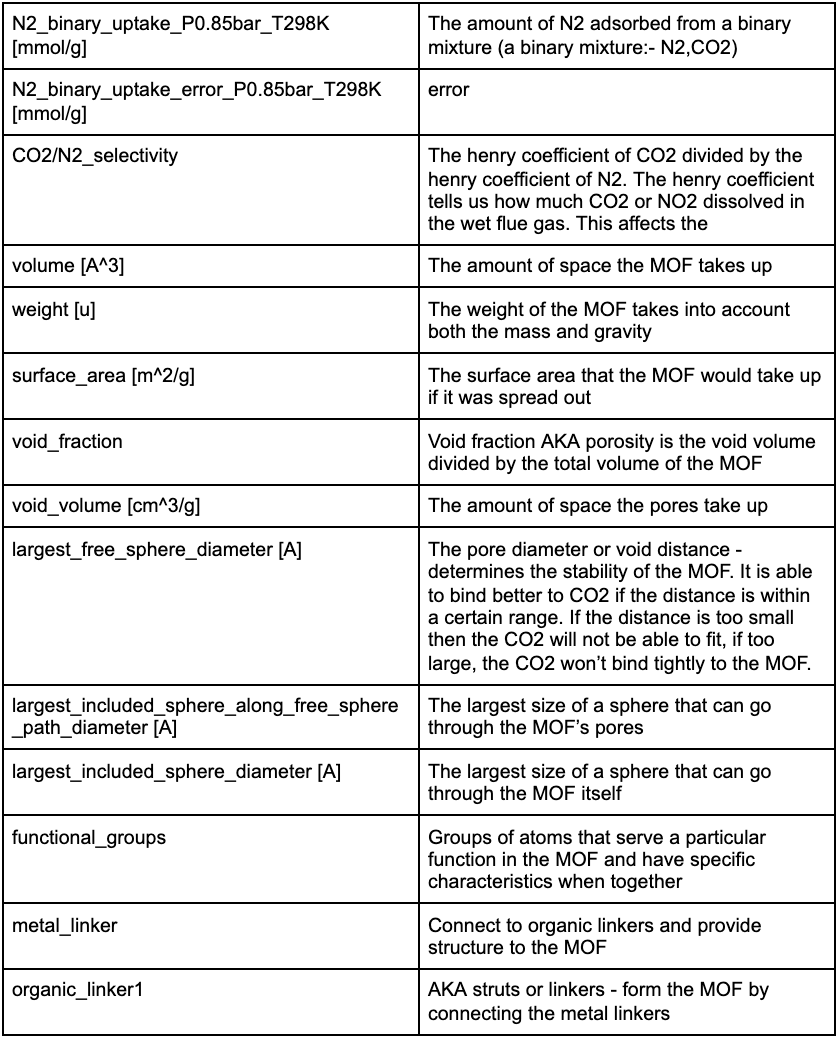}
    \includegraphics[width=\linewidth]{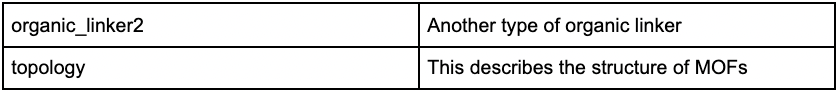}
    \caption{Descriptions of the different features used in the data set.}
    \vspace{-2.5mm}
    \label{fig:DataAnalysis}
\end{figure}

\section{OBSERVATIONS}
\label{sec:obs}

\subsection{Data Processing}
\label{subsec:obs}

The MOF name contains information about the structure, number of metal ions, number of organic linkers 1 and 2, topological structure, symmetry, and metric space groups. An example of a MOF name is str\_m5\_o16\_o16\_sra\_sym.77. In this case, "str" indicates structure, "m5" describes that there are 5 metal units, the first "o16" shows that there are 16 organic linker 1s, the second "o16" shows that there are 16 organic linker 2s, "sra" indicates the type of cubic periodic net, "sym" indicates symmetry, and "77" is the number of metric space groups in this particular MOF. Additionally, some of the contributors to MOF Efficiency are the henry coefficient, void volume, void distance, void fraction, and topological data. \cite{xu_azeem_izhar_shah_binyamin_aslam_2020} The Henry coefficient provides information about the solubility of gasses in the wet flue gas. It is calculated by dividing the concentration of species in the aqua space by the partial pressure of species in the gas phase. Inside the MOFs, there are pores, also known as voids.  The void volume and void distance describe the carbon capture capabilities of MOFs’ voids. The void fraction is calculated by dividing the void volume from the total volume. ~\cite{deng_grunder_cordova_valente_furukawa_hmadeh_gandara_whalley_liu_asahina_etal_2012} Topological data showcases the structural properties of the MOFs. The different classifications for the topological data in the database describe the type of periodic nets being used. Periodic nets show how frameworks are arranged. ~\cite{delgado-friedrichs_foster_o_keeffe_proserpio_treacy_yaghi_2005} The independent features that will be more highly considered for this research are the henry coefficients, void fraction, void volume, topological data, and void distance. For example, pcu describes an octahedron arrangement. In order to predict $\text{CO}_{\text{2}}$ efficiency, multiple dependent features must be considered such as $\text{CO}_{\text{2}}$/$\text{N}_{\text{2}}$ Selectivity, $\text{CO}_{\text{2}}$ uptakes at different pressures and temperatures, working capacity, and heat adsorption at different pressures and temperatures. These dependent features will mostly be separately analyzed as scientists will benefit from $\text{CO}_{\text{2}}$-adsorbing MOFs with different capabilities. For example, stability from low head adsorption might be compromised in certain cases in order to increase $\text{CO}_{\text{2}}$ uptake. If these dependent features are lumped together into a single efficiency formula, the complexity of the individual dependent features in relation to each other may be lost. \cite{rosen_iyer_ray_yao_aspuru-guzik_gagliardi_notestein_snurr_2021}
\section{METHODS }
\label{sec:methods}

\vspace{-0.05in}
\begin{figure}[hbt!]
    \centering
    \includegraphics[width=\linewidth]{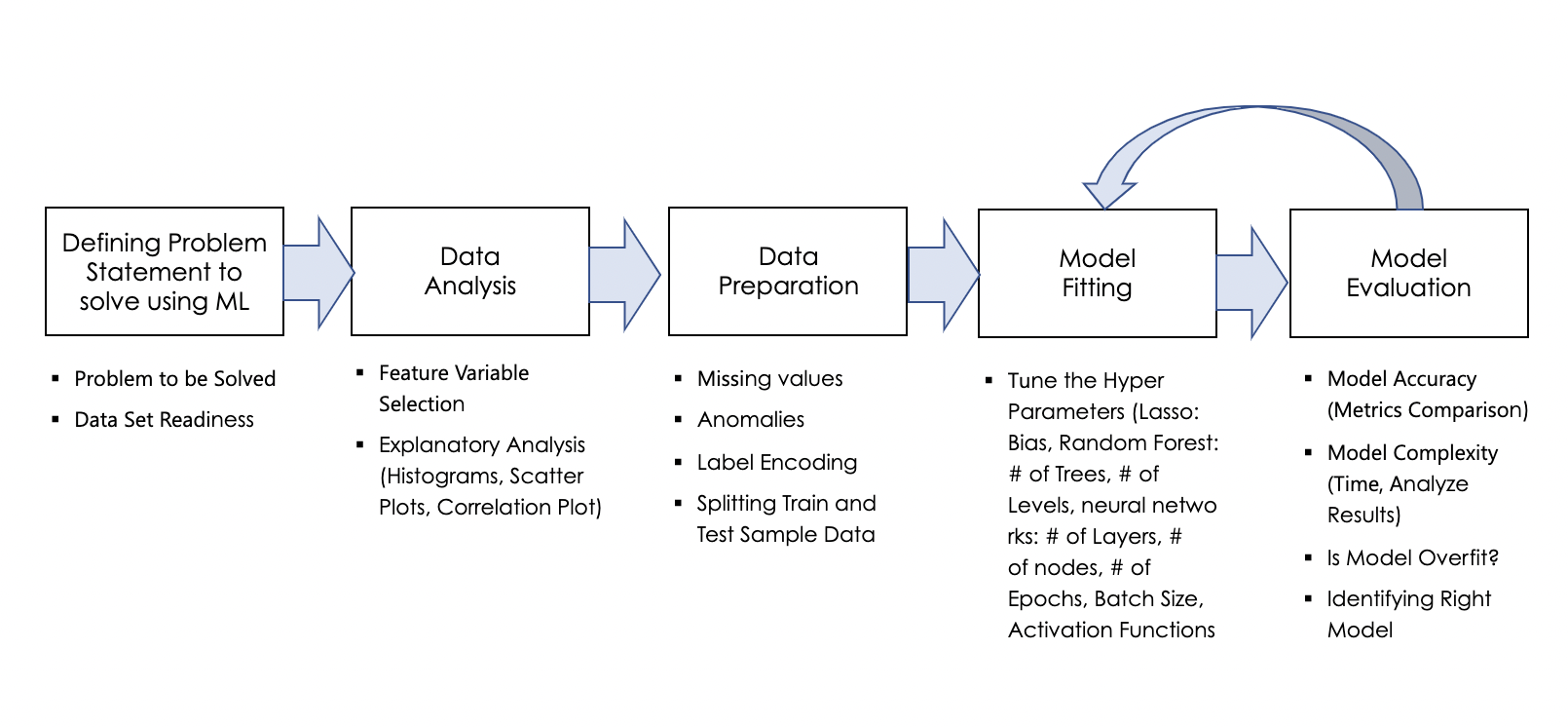}
    \caption{An overview of the process of the steps taken to create the Machine Learning models.}
    \vspace{-2.5mm}
    \label{fig:methods-1}
\end{figure}

\subsection{Data Pre-Processing}
\label{subsec:Data-Pre-Processing}

Before starting the project, it is imperative to understand and familiarize oneself to the correlations and trends present in the data set. The data set initially contained 324426 MOFs and 42 features. After cleaning the data of NaN values, null values, and anomalies, the result was that the data set contained 319290 MOFs and 15 features. In order to allow categorical variables to be observed by the machine learning algorithm, label encoding was utilized to assign a specific number to the different categories. This data was then standardized to the origin. After standardization, the independent and dependent variables were split. 70\% of the x and y became part of the training data and the remaining 30\% became part of the testing data. (Note: When creating graphs, it is important to use the non-standardized data so that negative values don't appear for features like volume or weight.) A correlation matrix was created in order to show the relationship between the different feature variables. Histograms were made in order to show where the data values were falling and what values were less common as well as what the anomalies were. One trend noticed was that the Carbon dioxide uptake and CO2/N2 selectivity showed a clear positive correlation. 

\subsection{Principal Component Analysis}
\label{subsec:Principal-Component-Analysis}

In the dataset, there were 42 different features. In order to improve the quality and decrease the complexity of the algorithm, this paper utilizes Principal Component Analysis, a method of dimensionality reduction. The process of PCA involves standardizing and re-plotting the original data points to create new principal components. In this way, the number of descriptors was reduced from 40 features to 7. The PCA graphs shown were only done for the  $\text{CO}_{\text{2}}$/$\text{N}_{\text{2}}$ Selectivity. \cite{ashutosh_tripathi_2019} 


\begin{figure}[h!]
    \centering
    \includegraphics[width=\linewidth, height=4cm]{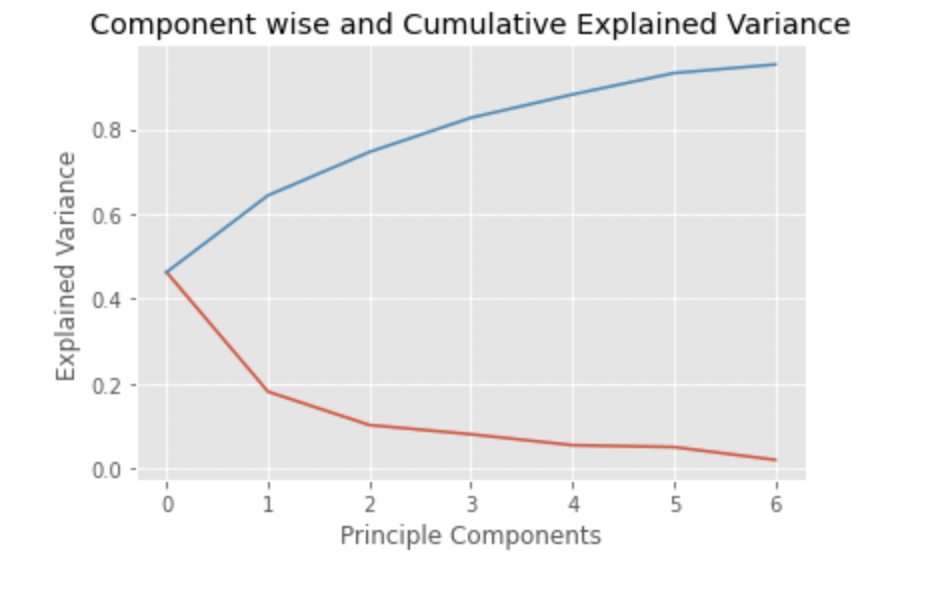}
    \caption{Number of principal components it takes to explain 90\% of the variance. The principal component analysis was able to reduce the dimensionality of the data from 40 to 7 principal components.}
    \label{fig:pca}
\end{figure}


\subsection{Linear, Lasso Regression}
\label{subsec:Linear-Regression}
In order to compare the descriptors and predict future MOF efficiencies in relation to those of the hypothetical MOFs, linear regression was used. The linear regression line transverses multiple dimensions and depicts the relationship between the dependent and independent features with a straight line. 
LASSO Regression is used to minimize overfitting which can occur due to having a large dataset. It uses a straight line to show the relationship between dependent and independent features like Linear Regression, yet a difference is that it adds biases to the data so that the overall line can be less fitted to certain points. Therefore, certain outliers have a lower effect on the general trends.The features without a coefficient of 0.00 were used for the LASSO Regression Model. 



\begin{figure}[h!]
    \centering
    \includegraphics[width=\linewidth,height=4cm]{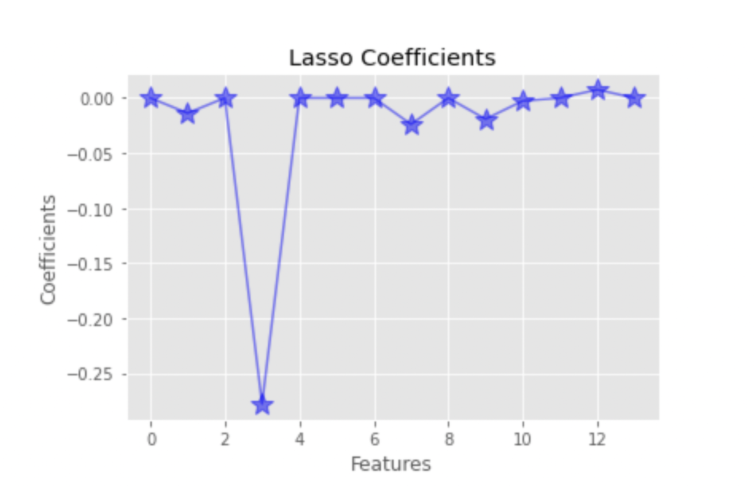}
    \caption{7 Features were deemed important in regards to this model.}
    \label{fig:Lasso}
\end{figure}



\subsection{Random Forest}
\label{subsec:Random-Forest}
Random Forests or Random Regressor Models create Decision Trees. These decision trees are constructed using the gini index which determines the descriptors’ importance and where they should be placed in the decision tree. 


\vspace{-0.05in}
\begin{figure}[hbt!]
    \centering
    \includegraphics[width=\linewidth]{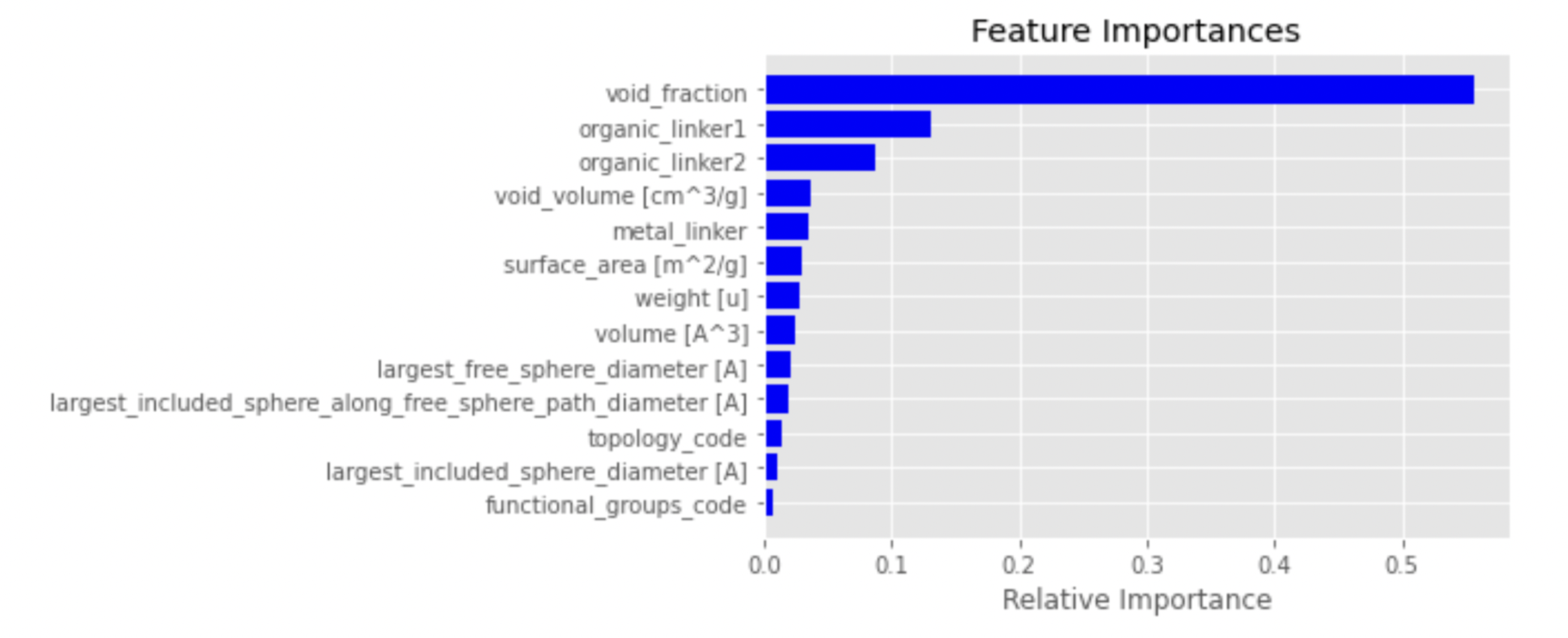}
    \caption{Feature Importance Chart. This shows the most important features that were used for the random forest in order with the coefficient or percentage of their importance.}
    \vspace{-2.5mm}
    \label{fig:RandomForest}
\end{figure}

\vspace{-0.05in}
\begin{figure}[hbt!]
    \centering
    \includegraphics[width=\linewidth]{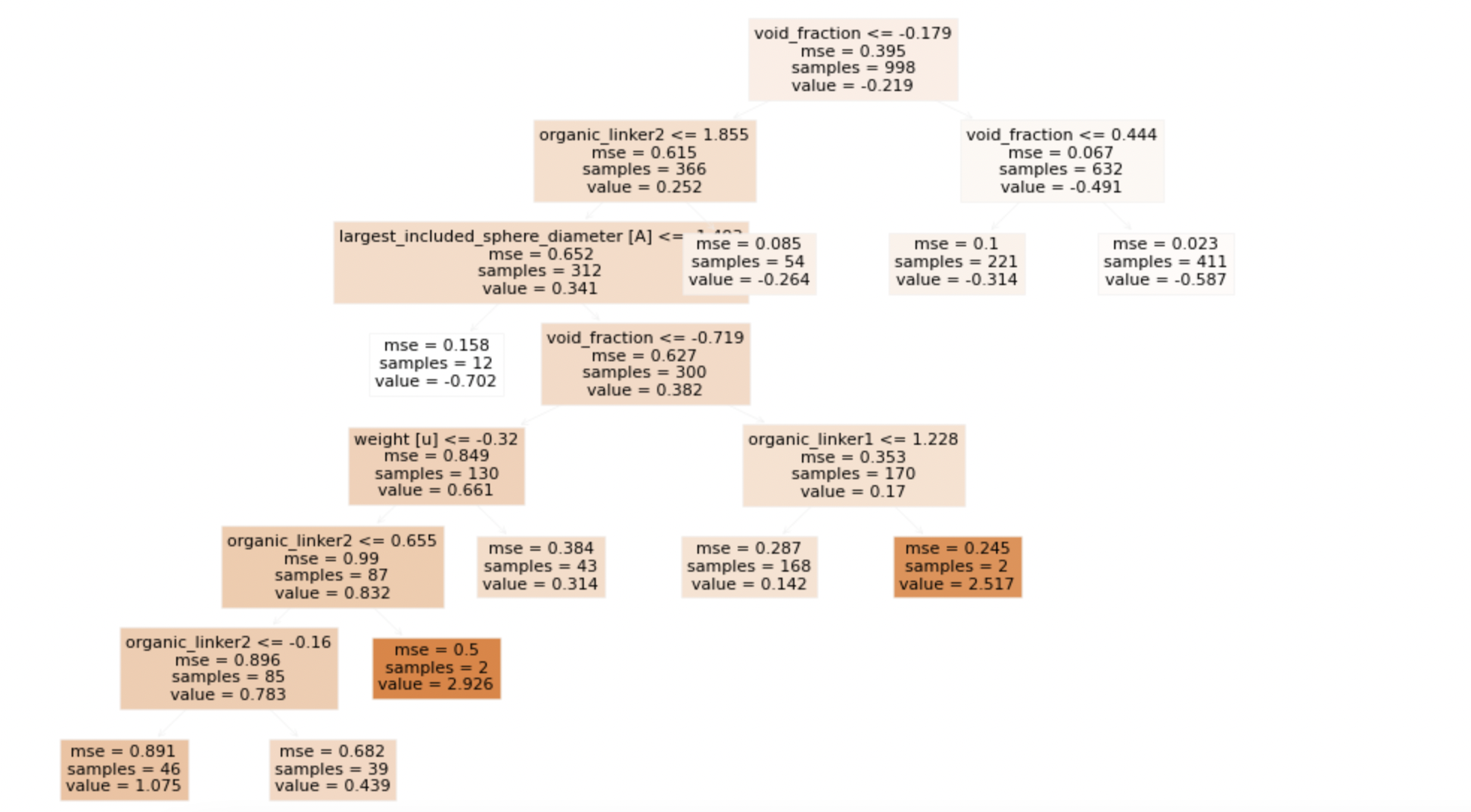}
    \caption{Random Forest Mapping. This describes the most important features selected by the Random Forest and the weights applied.}
    \vspace{-2.5mm}
    \label{fig:Random-Forest-Mapping}
\end{figure}


\subsection{Neural Networks}
\label{subsec:neural_networks}
Neural Networks are a network of input, hidden, and output layers that have inter-connected nodes. Initially, parameters are undefined and backpropagation is utilized to assign estimated values to them and minimize error. In order to fit the data, there exist activation functions such as RelU, Sigmoid, and Softplus. In this paper, RelU, and Linear are the activation functions being primarily used. 

\section{ML ALGORITHM ANALYSIS AND COMPARISON}
\label{sec:mla}
Based on the various algorithms used, it is evident that Dense Neural Networks was the most efficient in predicting the Carbon Dioxide Uptake of the MOFs in the data set. The Random Forest was the next most efficient. The model with the minimum mean squared error was recognized as the most efficient.

\subsection{Linear Regression, Linear Regression with principal component analysis, Lasso Regression}
\label{subsec:lrlasso}
Linear Regression and Linear Regression with principal component analysis had about 12\% error and were identical in their results. This outcome suggests that reducing dimensionality did not have any effect on the model's fit to the training data. Linear Regression did not result in high performance because the $\text{CO}_{\text{2}}$ uptake lacked a strong linear correlation with the input features. Despite testing with different alpha values, Lasso Regression had the most error when fitting to the train data. This result might have arisen due to a similar reason as the Linear Regression: even as biases were applied to the data, high variability resulting from a large data set might have led to the increased error. 

\subsection{Random Forest}
\label{subsec:RANDOMFOREST}
 The Random Forest's method of selecting and ranking features based on the Gini Index led the model to disregard most outliers and non-essential features which decreased error. Depending on the number of trees generated, the accuracy of the Random Forest could be more accurate, yet time taken would increase. The balance that was found in this paper was a forest depth of 10 levels. The random forest also provided a feature importance table (Figure 5), in which void fraction, number of organic linker 1 and 2, void volume, metal linker, and surface area were attributed to predicting the CO2 uptake. 

\subsection{Dense Neural Networks}
\label{subsec:DNN}
The Dense Neural Networks applied numerous weights and biases in multiple combinations of the training data set which led to accurate results for the testing data. The result was that the Neural Networks only had about 3\% error in predicting $\text{CO}_{\text{2}}$ uptake. This shows that if scientists apply Dense Neural Networks to their hypothetical data set, they will have low error when deciding which $\text{CO}_{\text{2}}$-adsorbing MOFs to synthesize. Figure 7 gives a comprehensive overview of the parameters and mean squared errors for the models used. Further tuning of hyper-parameters by changing the number of epochs or hidden layers may improve accuracy.

\vspace{-0.05in}
\begin{figure}[hbt!]
    \centering
    \includegraphics[width=\linewidth]{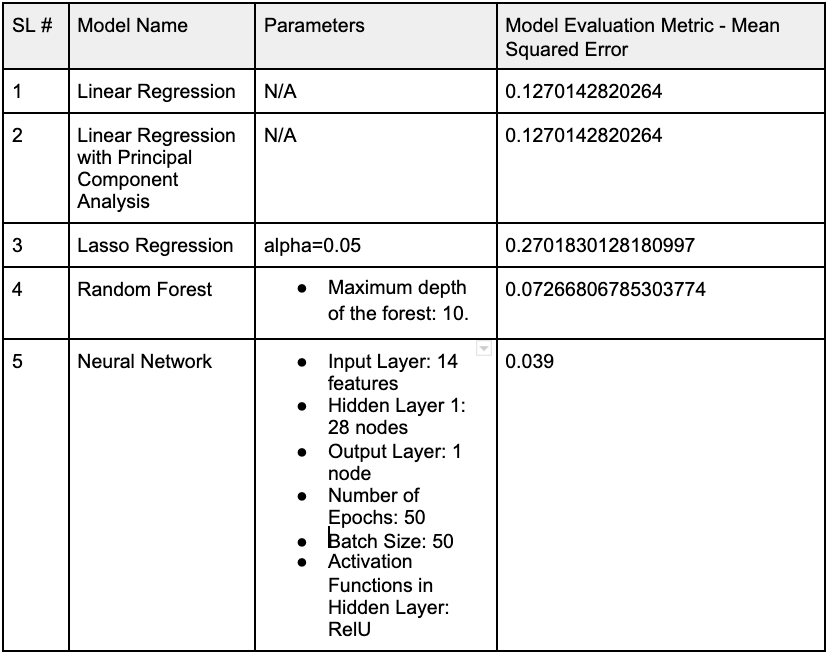}
    \caption{Results: Comparison of the different models..}
    \vspace{-2.5mm}
    \label{fig:ModelComparison}
\end{figure}

\section{LIMITATIONS AND FUTURE DIRECTIONS}
\label{sec:limitations}
Some limitations that were present when creating the models were for some of the MOFs the Carbon uptake was inversely proportional to certain feature features when the opposite relation is generally expected. For example, occasionally the volume of a certain MOF might be negative. The downfalls of using the hypothetical database is that the results could be skewed in favor of a theoretical situation. For example, a hypothetical MOF may show a certain $\text{CO}_{\text{2}}$ uptake, but due to inefficiencies in the real world, once synthesized, the MOF could adsorb a lower amount of $\text{CO}_{\text{2}}$. \cite{danaci_bui_mac_dowell_petit_2020} Certain correlations were not accurate and will be investigated. In the future, the development of a web application which uses the methods applied in this paper could be beneficial for chemists who may not be familiar with the technicalities of Machine Learning, but are interested in finding the predicted efficiency of their MOFs.

For further research, it would be important to test the Machine Learning models on other databases, specifically the CoRE MOF Database provided by Cambridge University. Furthermore, structural data through the format of cif files rather than just the names of the periodic nets from the topological data could have served to increase the efficiency of the ML models. This database would be beneficial to use as the data would be more realistic as it comes from previously synthesized MOFs. Other dependent features such as $\text{CO}_{\text{2}}$/$\text{N}_{\text{2}}$ Selectivity, and working capacity would be tested in accordance with $\text{CO}_{\text{2}}$ uptake to give a more holistic perception of the overall efficiency of the MOF.~\cite{tiotiu_novakova_nedeva_chong-neto_novakova_steiropoulos_kowal_2020} 

\section{SOURCE CODE}
\label{sec:sourcecode}
\url{https://github.com/mahu0926/MOF_ML}

\section{Acknowledgements}
\label{sec:acks}
I would like to show my appreciation for the following people who advised me during my research. Dr. Hao Zhuang introduced me to Metal-Organic Frameworks and provided supplementary articles which guided me. Christine mentored and provided resources for the machine learning techniques used. Dr. Ekashmi Rathore aided me with understanding feature variables in the MOF data. My advisors during this process displayed unfathomable patience and kindness, and encouraged me to complete my paper within the limited time frame. 

\bibliographystyle{format/IEEEtran}
\bibliography{References.bib}

\end{document}